\documentclass[10pt,twocolumn,letterpaper]{article}

\usepackage{iccv}              
\usepackage{comment}
\usepackage{amsmath, amssymb}   
\usepackage{graphicx}           
\usepackage{color}              
\usepackage{pifont}             
\usepackage{booktabs}           
\usepackage{multirow}           

\usepackage{times}  
\usepackage{helvet}  
\usepackage{courier}  
\usepackage[hyphens]{url}  
\usepackage{natbib}  
\usepackage{caption} 

\usepackage{algorithm}
\usepackage{algorithmic}
\usepackage{booktabs}
\usepackage{mathtools}
\usepackage{xcolor}
\usepackage{float}

\usepackage{pifont}%
\newcommand{\cmark}{\textcolor{teal}{\ding{51}}}%
\newcommand{\xmark}{\textcolor{red}{\ding{55}}}%
\newcommand{\ours}{\texttt{CHROME}}

\usepackage{graphicx}
\usepackage{multirow}
\usepackage{wrapfig}

\definecolor{iccvblue}{rgb}{0.21,0.49,0.74}
\usepackage[pagebackref,breaklinks,colorlinks,allcolors=iccvblue]{hyperref}

\newcommand\blfootnote[1]{%
  \begingroup
  \renewcommand\thefootnote{}\footnote{#1}%
  \addtocounter{footnote}{-1}%
  \endgroup
}

\def\paperID{694} 
\def\confName{ICCV}
\def\confYear{2025}

\title{CHROME: Clothed Human Reconstruction with Occlusion-Resilience and Multiview-Consistency from a Single Image}

\author{Arindam Dutta$^{1}$$^{\dagger}$ \ Meng Zheng$^{2}$ \ Zhongpai Gao$^{2}$ \ Benjamin Planche$^{2}$ \ Anwesa Choudhuri$^{2}$ \\ \ Terrence Chen$^{2}$ \ Amit K. Roy-Chowdhury$^{1}$ \ Ziyan Wu$^{2}$ \\
$^{1}$University of California, Riverside, CA $ \quad \quad ^{2}$United Imaging Intelligence, Boston, MA \\
{\tt\small \{adutt020@, amitrc@ece.\}ucr.edu,  \{first.last\}@uii-ai.com} 
}

\begin{document}

\twocolumn[{%
\renewcommand\twocolumn[1][]{#1}%
\maketitle
\begin{center}
    \centering
    \includegraphics[width=1\linewidth]{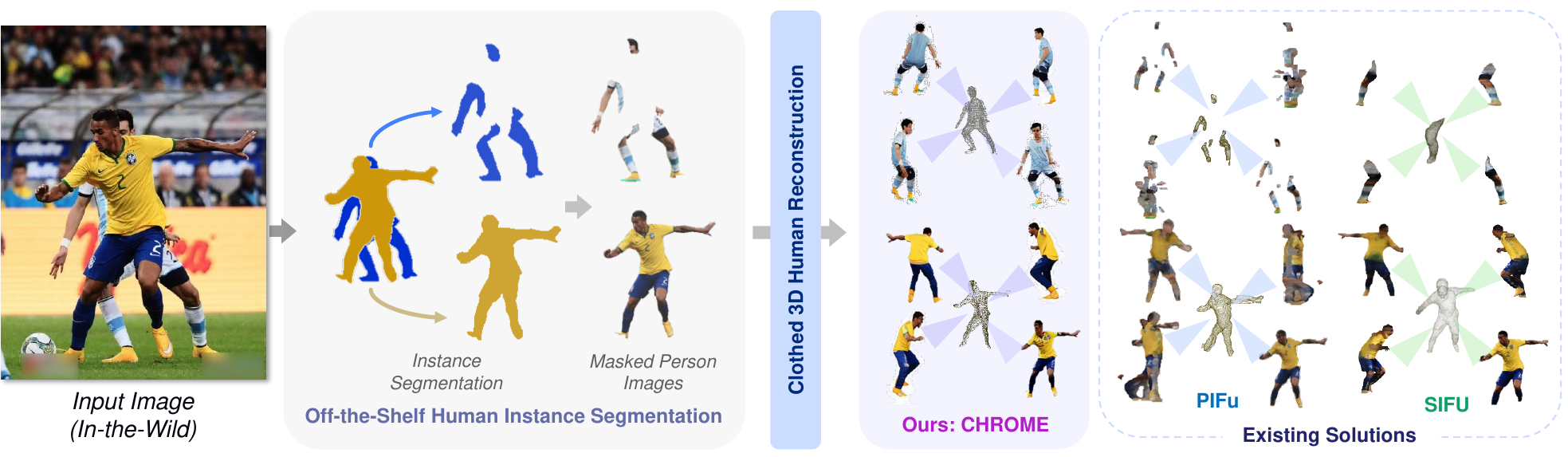}
    \captionof{figure}{\textbf{Need for occlusion-resilient multiview-consistent clothed human reconstruction from single images:} Existing algorithms for monocular 3D clothed human reconstruction, such as PIFu~\cite{saito2019pifu} and SIFU~\cite{zhang2024sifu}, produce fragmented and multiview-inconsistent novel view reconstructions from single-view occluded human images~\cite{zhang2019pose2seg}. In contrast, \ours~generates occlusion-free, multiview-consistent novel views from a single occluded image, using a novel pose-controlled multiview diffusion model, combined with a large reconstruction model, to create a cohesive 3D representation of the human subject. Moreover, \ours~does not require 3D mesh supervision during training as required for existing algorithms~\cite{saito2019pifu, zhang2024sifu}. 
    }
    \label{fig:teaser_1}
    
\end{center}%
}]

\blfootnote{\noindent $\dagger$ This work was done during Arindam Dutta's internship at United Imaging Intelligence, Boston, MA.}

\begin{abstract}
Reconstructing clothed humans from a single image is a fundamental task in computer vision with wide-ranging applications. Although existing monocular clothed human reconstruction solutions have shown promising results, they often rely on the assumption that the human subject is in an occlusion-free environment. Thus, when encountering in-the-wild occluded images, these algorithms produce multiview inconsistent and fragmented reconstructions. Additionally, most algorithms for monocular 3D human reconstruction leverage geometric priors such as SMPL annotations for training and inference, which are extremely challenging to acquire in real-world applications.To address these limitations, we propose $\ours$: \underline{C}lothed \underline{H}uman \underline{R}econstruction with \underline{O}cclusion-Resilience and \underline{M}ultiview-Consist\underline{E}ncy from a Single Image, a novel pipeline designed to reconstruct occlusion-resilient 3D humans with multiview consistency from a single occluded image, without requiring either ground-truth geometric prior annotations or 3D supervision. Specifically, $\ours$ leverages a multiview diffusion model to first synthesize occlusion-free human images from the occluded input, compatible with off-the-shelf pose control to explicitly enforce cross-view consistency during synthesis.
A 3D reconstruction model is then trained to predict a set of 3D Gaussians conditioned on both the occluded input and synthesized views, aligning cross-view details to produce a cohesive and accurate 3D representation. $\ours$ achieves significant improvements in terms of both novel view synthesis (upto 3 db PSNR) and geometric reconstruction under challenging conditions. 
\end{abstract}

\vspace{-0.5cm}


\section{Introduction}
The task of reconstructing a 3D clothed human reconstruction involves generating a 3D human model from multiview images~\cite{zheng2024gps} or monocular videos~\cite{weng2022humannerf} with numerous applications spanning augmented reality, virtual try-on and digital avatars~\cite{minar20203d, zhang2023getavatar}. However, these approaches are often impractical as capturing multiview images in real-world settings is neither budget-friendly nor scalable. This limitation has driven researchers to focus on \emph{single-view 3D clothed human reconstruction} algorithms, which encompasses reconstructing a 3D human model from just a single image~\cite{saito2019pifu, saito2020pifuhd, zhang2024global, zhang2024sifu, ho2024sith}. 
However, 
\emph{occlusions remain a major obstacle, often resulting in fragmented and inaccurate multi-view reconstructions}~\cite{zhou2021human, sun2024occfusion}, making existing approaches impractical for real world settings. 
Although recent studies have improved occlusion resilience for specific low-level human modeling tasks such as segmentation~\cite{dutta2024poise} and pose estimation~\cite{liu2022view, zhang20233d}, \textbf{reconstructing occlusion-resilient and multi-view consistent} texture and 3D geometric details from novel camera views under occlusion remains largely unaddressed. 
As visualized in Figure~\ref{fig:teaser_1}, existing algorithms for single-view 3D clothed human reconstruction~\cite{saito2019pifu, zhang2024sifu} perform poorly in in-the-wild scenarios when the human subject is partially occluded, resulting in inconsistent and fragmented novel view synthesis. 

Existing 3D clothed human reconstruction methods can be divided into several main streams based on different volumetric representations, traditional 3D human reconstruction learns parametric models such as SMPL~\cite{loper2023smpl}, which show moderate occlusion resilience by embedding human body priors but prioritize surface reconstruction over texture~\cite{xiu2022icon, xiu2023econ}, with limited appearance detail for clothed human reconstruction. However, these methods typically require large-scale image and SMPL pose/shape pairs for training, which demands labor-intensive annotation and often compromises generalizability under in-the-wild scenes. 
On the other hand, implicit representations like Neural Radiance Fields (NeRFs)~\cite{mildenhall2021nerf, weng2022humannerf} and point-based representations like 3D Gaussian Splatting (3DGS)~\cite{kerbl3Dgaussians, hu2024gauhuman} offer high visual-fidelity human reconstruction from monocular videos. However, they struggle with occlusions as they often require pixel-level fine details for subject-specific optimization, which can be largely affected by occlusion noises, 
as discussed in~\cite{xiang2023rendering, sun2024occfusion}. 

To address these challenges, we propose \textbf{$\ours$}: \emph{\underline{C}lothed \underline{H}uman \underline{R}econstruction with \underline{O}cclusion-Resilience and \underline{M}ultiview-Consist\underline{E}ncy}, 
which leverages a multiview diffusion model to generate consistent, occlusion-free multiview human images to learn a 3D Gaussian representation from occluded monocular image input. 
Specifically $\ours$ adopts a two-stage learning strategy. In the first stage, $\ours$ addresses the limited information available in a single-view occluded image using a multiview diffusion model~\cite{shi2023zero123++}, compatible with off-the-shelf pose control, to generate four occlusion-free cross-view consistent human images. 
Using an estimated 3D pose as an explicit guidance signal, we ensure enhanced pose consistency across the generated views, which is essential for a coherent 3D human representation. The second stage of $\ours$ incorporates a 3D reconstruction model that learns to predict a set of 3D Gaussians~\cite{kerbl3Dgaussians}, providing a cohesive 3D representation of the human subject. This model is conditioned on both the initial occluded image and the synthesized multiview occlusion-free images, enabling it to capture and integrate geometric and texture details consistently across views. 

The proposed two-stage approach effectively bridges 2D-3D information gaps, advancing the capability of single-image-based 3D reconstruction under occlusions. 
On the other hand, by optimizing the photometric loss in the 2D projected space $\ours$ is able to take advantage of large-scale 2D human images for implicit geometric prior learning, eliminating the need to require 3D supervision and SMPL priors as in existing solutions~\cite{saito2019pifu,ho2024sith,zhang2024sifu, zhang2024global, zheng2021pamir}, resulting in superior generalizability under occlusions in real-world application scenarios.


\noindent The following are the main \underline{contributions} of the work.
\begin{itemize}
    \item We address the problem of multiview reconstruction of a human subject under occlusions from a single view image. This addresses the bottlenecks associated with existing algorithms, which perform poorly under occlusions. 
    \item To tackle the aforementioned problem, we propose $\ours$, a novel algorithm that utilizes a multiview diffusion model, enabling off-the-shelf pose control to generate cross-view de-occluded images from a single occluded input. A reconstruction model then predicts a cohesive set of 3D Gaussians conditioned on both the occluded input and the synthesized views, enabling occlusion-free novel view synthesis. 
    \item $\ours$ does not require ground-truth SMPL mesh annotations as in existing algorithms. We further demonstrate the generalizability of the proposed pipeline with stereo input with improved reconstruction accuracy and cross-view consistency. 
    \item Extensive experiments showcase the strength of $\ours$ in novel view synthesis from a single occluded human image in both in-domain and zero-shot settings.
\end{itemize}

\section{Related Works}

\noindent \textbf{Monocular Human Mesh Recovery:} Monocular Human Mesh Recovery aims to generate a 3D human mesh from a single image~\cite{kanazawa2018end}, predominantly using the SMPL model~\cite{bogo2016keep} and its variants for 3D pose and shape estimation. Recent advances~\cite{joo2021exemplar, moon2020i2l, kocabas2021pare, yoshiyasu2023deformable} have leveraged large-scale synthetic~\cite{black2023bedlam} and real-world datasets~\cite{lin2014coco, andriluka2014mpiinf, joo2015panoptic}, improving reconstruction for diverse datasets and in-the-wild scenarios~\cite{goel2023humans}. However, challenges remain in handling occlusions, with recent methods~\cite{khirodkar2022occluded, zhang2020object, chen2023mhentropy} focusing on occlusion-robust algorithms often using learned priors like~\cite{tiwari2022pose, ta2024multi} to enhance reconstruction reliability. Despite these advancements, these algorithms focus primarily on bare-body mesh recovery, often overlooking the challenges of reconstructing texture and clothing details.

\noindent \textbf{Clothed Human Reconstruction:} Clothed human reconstruction focuses on generating a 3D mesh of clothed subjects, textured or untextured, from various inputs such as single images, multiview images, or monocular videos. Existing methods fall primarily into two categories: implicit function-based techniques~\cite{saito2019pifu, xiu2022icon, zheng2021pamir, xiu2023econ, zhang2024global, ho2024sith, zhang2024sifu} and rendering-based approaches~\cite{weng2022humannerf, hu2024gauhuman, hu2023sherf, pan2024humansplat, chen2024generalizable}. Implicit function-based methods, such as PIFu~\cite{saito2019pifu} and its extensions~\cite{saito2020pifuhd}, leverage pixel-aligned features processed through learnable decoders to reconstruct detailed 3D models using 3D supervision. Recent approaches like GTA~\cite{zhang2024global} and SIFU~\cite{zhang2024sifu} incorporate the SMPL model to embed anatomical priors, improving the fidelity of reconstruction. TeCH~\cite{huang2024tech} applies optimization techniques to achieve single-image 3D digitization. However, occlusions persist as a critical challenge in clothed human reconstruction~\cite{mhcdiff2024}. Wang \textit{et al.}~\cite{wang2023complete} proposed a methodology utilizing ground-truth SMPL meshes to mitigate occlusion effects. However, their reliance on precise SMPL mesh estimations, which are inherently challenging to acquire under occluded conditions obstruct practical adoption. With advances in Neural Radiance Fields (NeRFs)~\cite{mildenhall2021nerf} and 3D Gaussian Splatting (3DGS)~\cite{kerbl3Dgaussians}, studies such as~\cite{weng2022humannerf, hu2024gauhuman} have adapted these techniques for human reconstruction from monocular videos, though occlusion handling remains problematic, as evidenced in occlusion-robust variants~\cite{xiang2023rendering, xiang2023wild2avatar, sun2024occfusion}. Despite this progress, they typically require monocular video inputs and are optimized per subject, limiting their scalability and practical deployment. Differently, we propose a single image-based generalizable approach for occlusion-free novel view synthesis and geometric reconstruction.

\noindent \textbf{Multi-view Diffusion and Large Reconstruction Models for 3D Reconstruction:} Latent Diffusion Models (LDMs)~\cite{rombach2022high} have aided significant advancements in image generation, reconstruction, and single-image-to-3D tasks. Zero-1-to-3~\cite{liu2023zero} pioneered the use of LDMs to generate novel views from single input images with specified camera parameters, sparking a wave of related research~\cite{shi2023zero123++, shi2023mvdream, gao2024cat3d, qian2023magic123, voleti2025sv3d}. ControlNet~\cite{zhang2023adding} introduced guided control over LDM sampling, advancing conditional generation. Recent works facilitate the creation of 3D objects from text prompts~\cite{ju2023humansd, lu2024direct2} by leveraging extensive training datasets~\cite{schuhmann2022laion}. Pretrained LDMs also function as priors for novel view synthesis, modeling natural image distributions, and penalizing the reconstruction model through Score Distillation Sampling based losses~\cite{poole2022dreamfusion, gu2023nerfdiff, lee2024guess}. Recently, the Large Reconstruction Model (LRM)~\cite{hong2023lrm} introduced NeRF-based triplane representations for efficient single image to 3D generation, trained on large multiview datasets~\cite{objaverse, objaverseXL}. Large Gaussian Model (LGM)~\cite{tang2024lgm} extended the same by integrating pretrained diffusion models for single-image-to-3D synthesis, achieving faster rendering with 3DGS techniques. However, these algorithms struggle with partial occlusions, motivating our proposed approach for 3D human recovery from single or stereo occluded images.
\begin{figure*}[!t]
    \centering
    \includegraphics[width=0.9\textwidth]{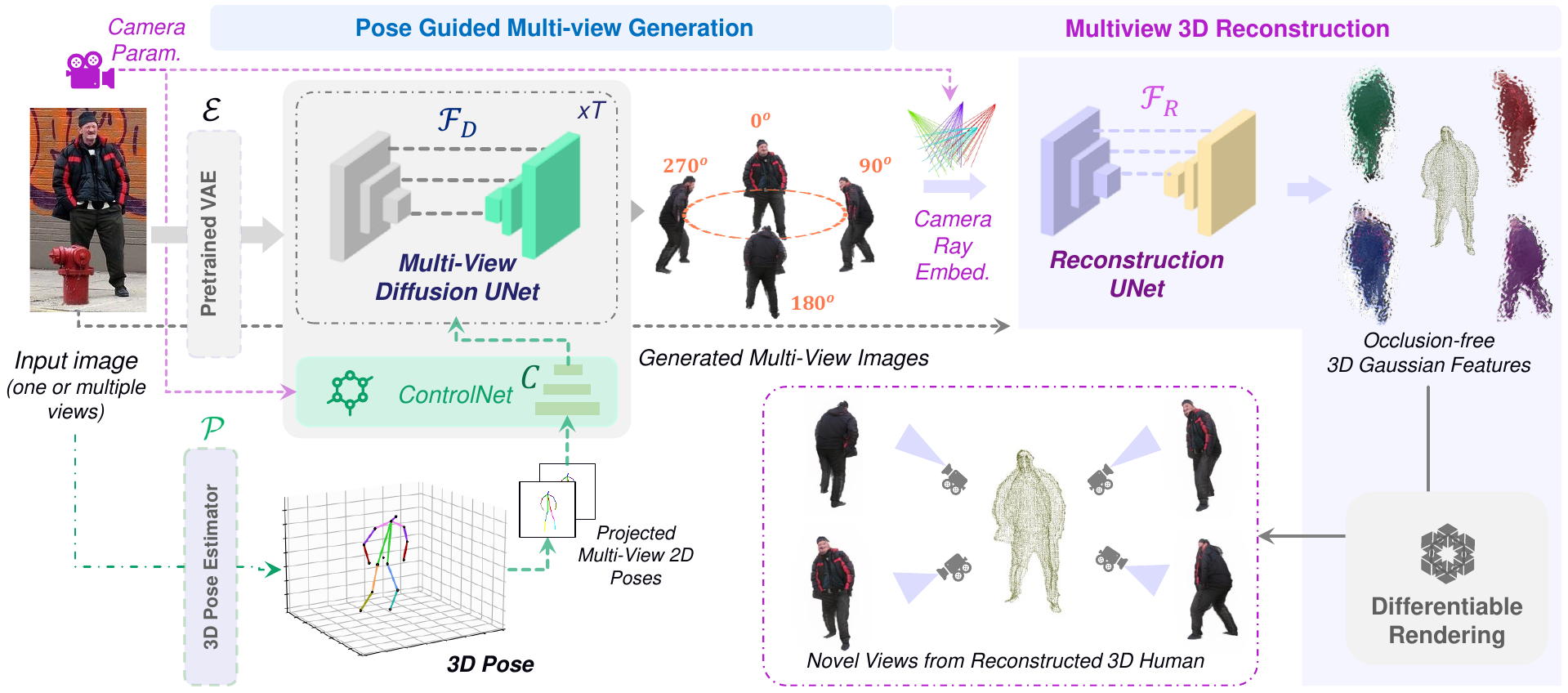}
    \caption{\textbf{Overview of proposed method:} $\ours$ is a novel two-stage pipeline designed for occlusion-free 3D human reconstruction from a single occluded image. In the first stage, a pose-controlled multiview diffusion model generates four de-occluded views of the subject, ensuring pose consistency across the synthesized images, conditioned on the input occluded image and the 3D pose estimates. In the second stage, a 3D  reconstruction model combines the occluded input image with the synthesized multi-view images to create a cohesive 3D Gaussian representation, aligning geometric and texture details across views, enabling accurate reconstructions and robust novel view synthesis, even in the presence of significant occlusions. Note that the method is also compatible with  multi-view inputs.}
    \label{fig:pipeline} 
    \vspace{-1em}
\end{figure*}

\section{Methodology}

In this section, we introduce our proposed algorithm $\ours$, which is designed to generate occlusion-free novel views of a human subject from a single image, where the subject can be partially obscured by occlusions. Specifically, given an input image \( x \in \mathbb{R}^{H \times W \times 3} \), representing a partially visible human subject within spatial dimensions \( H \) and \( W \), $\ours$ reconstructs \( N \) occlusion-free novel views (denoted as \( y \in \mathbb{R}^{N \times H \times W \times 3} \)) of this subject. Additionally, while we use a single-view image as input for clearer illustration, it is important to note that our pipeline is fully compatible with multiview inputs.

The pipeline $\ours$ is structured with two distinct modules and is trained from end-to-end to achieve novel view synthesis from a single occluded image \( x \). The first module, \( \mathcal{F}_{D} \), leverages the power of latent diffusion models~\cite{rombach2022high} in conjunction with ControlNet~\cite{zhang2023adding} to synthesize four pose-controlled, multiview-consistent images from the occluded input. These occlusion-free images form the basis for novel synthesis under occluded single-view inputs. The second module, $\mathcal{F}_{R}$, learns to reconstruct occlusion-free views by combining information from both the original occluded input and the  $\mathcal{F}_{D}$ generated images. This results in \( N \) occlusion-free novel views of the human subject.
\begin{equation}
    y = \mathcal{F}_{R} \circ \mathcal{F}_{D} (x),
\end{equation}
where the composition of functions \( \mathcal{F}_{R} \) and \( \mathcal{F}_{D} \) denotes the sequential application of the diffusion and reconstruction modules. The complete architecture of the proposed algorithm, $\ours$, is depicted in Figure~\ref{fig:pipeline}.

\subsection{Pose Controlled Multiview Consistent Diffusion}

Given a single occluded image \( x \), the objective of \( \mathcal{F}_{D} \) is to generate four pose-controlled, multiview consistent images, represented by \( y_{D} \in \mathbb{R}^{4 \times H \times W \times 3} \). Each of these images in \( y_{D} \) corresponds to a unique view, achieved by rotating the camera around a fixed radius at angles of \( 0^{\circ} \), \( 90^{\circ} \), \( 180^{\circ} \), and \( 270^{\circ} \). We employ a latent diffusion model (LDM)~\cite{rombach2022high} to define $\mathcal{F}_{D}$, allowing it to learn the conditional distribution of these four multiview images based on the occluded input image. The denoising diffusion process operates within the latent space of $\mathcal{F}_{D}$ leveraging a pretrained variational autoencoder (VAE) that efficiently encodes and decodes latent vectors for image reconstruction. 

The task of reconstructing multiview images based on a single occluded image is inherently ill-posed and underconstrained, making it essential to introduce suitable conditions to regularize the learning process. To address this, we employ a pre-trained VAE encoder module, $\mathcal{E}$, to extract features from the visible regions of $x$, which are then passed on to $\mathcal{F}_{D}$. This step ensures conditioning the multiview generation on the observable portions of $x$, ensuring that the output images maintain a consistent appearance with the visible parts of the occluded input image. However, we observe that multiview image generation with conditioning $\mathcal{F}_{D}$ solely on occluded input image features sometimes results in multiview images with suboptimal cross-view consistency. Hence, we utilize a ControlNet~\cite{zhang2023adding} ($\mathcal{C}$) based approach to explicitly condition the diffusion process on the 2D target poses ($\mathbf{P}^{2d}_{D}$), ensuring pose consistency across generated multiview images. This strategy further regularizes $\mathcal{F}_{D}$, guiding it to produce consistent multiview pose-aligned output images. 

To train the model $\mathcal{F}_{D}$, we first generate paired input and multiview target images with camera orientations set at \( 0^{\circ} \), \( 90^{\circ} \), \( 180^{\circ} \), and \( 270^{\circ} \) relative to the camera parameters of the input image, using rendered images from $\mathcal{S}$. Following the approach of~\cite{zhao2024large}, random occlusions are applied to the input images, resulting in paired occluded input images $x$ and corresponding ground-truth multiview target images $y_{D}$. The ground-truth latent vector is defined as $z_{0} = \mathcal{E}(y_{D})$ and undergoes perturbation through the diffusion process across $t$ timesteps to produce $z_{t}$. The objective of $\mathcal{F}_{D}$ is to recover the perturbation noise $\epsilon_{\mathcal{F}_{D}}$ at each timestep $t$, conditioned on features from the occluded input image $\mathcal{E}(x)$ and a control signal provided by the ground-truth 2D target poses through ControlNet, denoted as $\mathcal{C}(\mathbf{P}^{2d}_{D})$. Thus, the training objective can be formally expressed as
\begin{equation}
    \min_{\mathcal{F}_{D}} \quad \mathbb{E}_{z \sim \mathcal{E}(x), t, \epsilon \sim \mathcal{N}(0, I)} \left\| \epsilon - \epsilon_{\mathcal{F}_{D}}(z_t; t, \mathcal{E}(x), \mathcal{C}(\mathbf{P}^{2d}_{D})) \right\|_2^2.
\end{equation}

Furthermore, since diffusion models generally require large-scale training data to achieve effective convergence, we follow prior work~\cite{he2022masked, ho2024sith} by initializing $\mathcal{F}_{D}$ with pretrained weights from \cite{shi2023zero123++}, which was trained on an extensive collection of multiview images from \cite{objaverse}. In our training, we fine-tune only the parameters of the UNet~\cite{rombach2022high} and ControlNet~\cite{zhang2023adding} models, leaving all other parameters, including the encoder and decoder of the VAE fixed. Furthermore, by conditioning $\mathcal{F}_{D}$ on the features of the input image $\mathcal{E}(x)$, we ensure that even under incorrect pose predictions, $y_{D}$ retains the information in the original input image $x$ (see the supplementary for more discussion).

\begin{figure}[!htb]
    \centering
    \includegraphics[trim={75 10 0 10}, clip, width=0.9\linewidth]{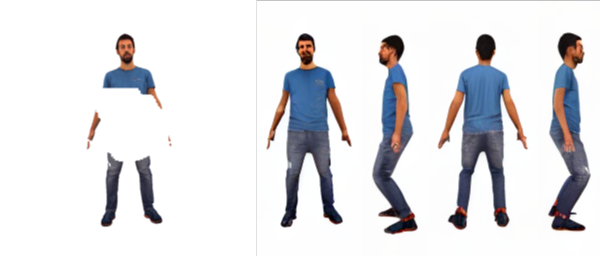}
    \caption{\textbf{Qualitative analysis of $\mathcal{F}_{D}$ in generating occlusion-free novel views:} Qualitative results demonstrating the capacity of $\mathcal{F}_{D}$ in generating occlusion-free novel views conditioned on a single occluded image from CustomHumans.
    }
    \label{fig:diff_main}
    \vspace{-0.4cm}
\end{figure}

During the inference phase, we employ a state-of-the-art 3D pose estimation model, $\mathcal{P}$~\cite{goel2023humans}, to estimate the 3D pose, denoted as $\hat{\mathbf{P}}^{3d}$, of the human subject from the occluded image input $x$. This extraction is represented by $\hat{\mathbf{P}}^{3d} = \mathcal{P}(x)$. Using a weak-perspective projection, we then obtain the corresponding 2D pose $\hat{\mathbf{P}}^{2d}_{D}$ for each multiview image. To generate the estimated multiview images $\hat{y}_{D}$, we utilize the pretrained VAE decoder module, $\mathcal{D}$, applied on the denoised latent vector $\hat{z}_{0}$ which is acquired through an iterative DDIM denoising (sampling) process~\cite{ho2020denoising} over $T$ timesteps, beginning from Gaussian noise and conditioned on both the occluded input image embedding $\mathcal{E}(x)$ and the estimated multiview 2D pose conditioning $\mathcal{C}(\hat{\mathbf{P}}^{2d}_{D})$. Mathematically, $\hat{y}_{D} = \mathcal{D}(\hat{z}_{0})$. Figure~\ref{fig:diff_main} shows the qualitative results of our $\mathcal{F}_{D}$ in generating occlusion free novel views conditioned on the occluded input image.

\subsection{Novel View Synthesis using Gaussian Splatting}

To achieve an occlusion-free 3D representation of the human subject using the four multiview images produced by our proposed pose-controlled multiview diffusion model, we build upon recent advancements in 3D Gaussian Splatting~\cite{kerbl3Dgaussians}. Specifically, our aim is to use the synthesized multiview images, generated by our diffusion function $\mathcal{F}_{D}$, alongside the original occluded input image, denoted \( x \), to inform a learning framework that maps this set of images to a 3D Gaussian representation. This mapping ($\mathcal{F}_{R}$) facilitates the transition from a spatially diverse image array to a continuous Gaussian field that approximates the underlying human subject in 3D space.

Formally, we define $\mathcal{F}_{R}$ as a transformation that operates on a tensor of five images, each of dimensions \( H \times W \) and containing RGB values, such that $\mathcal{F}_{R}: \mathbb{R}^{5 \times H \times W \times 3} \rightarrow \mathbb{R}^{n_{g} \times 14}$, where \( n_{g} \) denotes the number of Gaussians representing the human subject. Each Gaussian is represented by a precise set of parameters: mean \(\in \mathbb{R}^3 \), scaling \(\in \mathbb{R}^3 \), rotation quaternion \(\in \mathbb{R}^4 \),  opacity value \(\in \mathbb{R} \), and colors \( \in \mathbb{R}^3 \). Together, these elements constitute a 14-dimensional vector that encapsulates both spatial and appearance information essential for Gaussian representation. This compact representation supports efficient computation while preserving essential structural and visual details in the 3D space. By aligning these Gaussian representations across the multiview images, we achieve a cohesive occlusion-resilient 3D model of the underlying human subject.

To model the function $\mathcal{F}_{R}$, we employ an asymmetric UNet architecture~\cite{tang2024lgm} specifically designed to incorporate information from both the occluded input image and the synthesized multiview images. This architecture enables a seamless fusion of spatial and appearance details, ensuring a holistic 3D representation. Following the approach outlined in~\cite{xu2023dmv3d}, we encode camera poses into a 9-channel feature map, combining RGB values with ray embeddings to produce a unified representation that integrates color and spatial orientation for input to $\mathcal{F}_{R}$. Drawing inspiration from~\cite{szymanowicz2024splatter}, we treat each pixel in the resulting output feature map as an individual 3D Gaussian, forming a continuous Gaussian field that effectively captures both local details and spatial coherence. The final 3D Gaussian representation is achieved by concatenating the output Gaussians from each view, which are then rendered at the target resolution for supervision for generating multiview images.

To supervise the learning of $\mathcal{F}_{R}$, we utilize a differentiable renderer~\cite{kerbl3Dgaussians} that projects the final 3D Gaussian representation into a set of multiview images based on predefined camera parameters. Specifically, $N$ multiview images, denoted as $\{\hat{y}_{i}\}_{i=1}^{N}$, are generated from the final 3D Gaussian representation. These rendered images are then used to compute a combined loss function, encompassing photometric loss in the rendered RGB images and shape loss in the rendered $\alpha$ image~\cite{kerbl3Dgaussians} and the ground truth silhouettes $\{y_{silh}^{gt}\}_{i=1}^{N}$, with respect to the ground truth multiview images $\{y_{i}^{gt}\}_{i=1}^{N}$, which guide the optimization of $\mathcal{F}_{R}$. The objective function is formulated as follows:
\begin{equation}
    \min_{\mathcal{F}_{R}} \quad \mathcal{L}_{\text{mse}}(\hat{y}, y^{\text{gt}}) + \lambda_{1} \mathcal{L}_{\text{lpips}}(\hat{y}, y^{\text{gt}}) + \lambda_{2} \mathcal{L}_{\text{mse}}(\hat{y}_{\alpha}, y_{silh}^{\text{gt}})
    \label{eqn:obj}
\end{equation}

The weights $\lambda_{1}$ and $\lambda_{2}$ control the relative influence of each term in the objective function, and $\mathcal{L}_{\text{mse}}$ represents the mean squared error loss, $\mathcal{L}_{\text{lpips}}$ is the perceptual loss~\cite{zhang2018unreasonable}. Additionally, during \ours's end-to-end training, we enable \emph{estimated} 3D pose control derived from occluded images instead of ground-truth poses, which enhances \ours~'s generalizability and robustness in real-world scenarios where perfect pose information is not available. \\

\noindent To enable learning of $\mathcal{F}_{D}$ and $\mathcal{F}_{R}$, we utilize a dataset $\mathcal{S}$~\cite{tao2021function4d} consisting of textured human scans. For each scan, we render the corresponding images by positioning 16 equidistant perspective cameras at a fixed elevation, scene radius, and focal lengths, following the setup in~\cite{zheng2024gps}. During inference, given an occluded image \( x \), we first compute synthetic occlusion-free images as \( \hat{y}_{D} = \mathcal{F}_{D}(x) \). Next, the concatenated input \( \left[ x; \hat{y}_{D} \right] \) is passed through \( \mathcal{F}_{R} \) to generate \( N \) novel view images, denoted by \( \{ \hat{y}_{i} \}_{i=1}^{N} \), which can be mathematically expressed as $
    \{ \hat{y}_{i} \}_{i=1}^{N} = \mathcal{F}_{R}(\left[ x; \hat{y}_{D} \right])$

\begin{figure*}[!htb]
    \centering
    \includegraphics[width=0.85\linewidth]{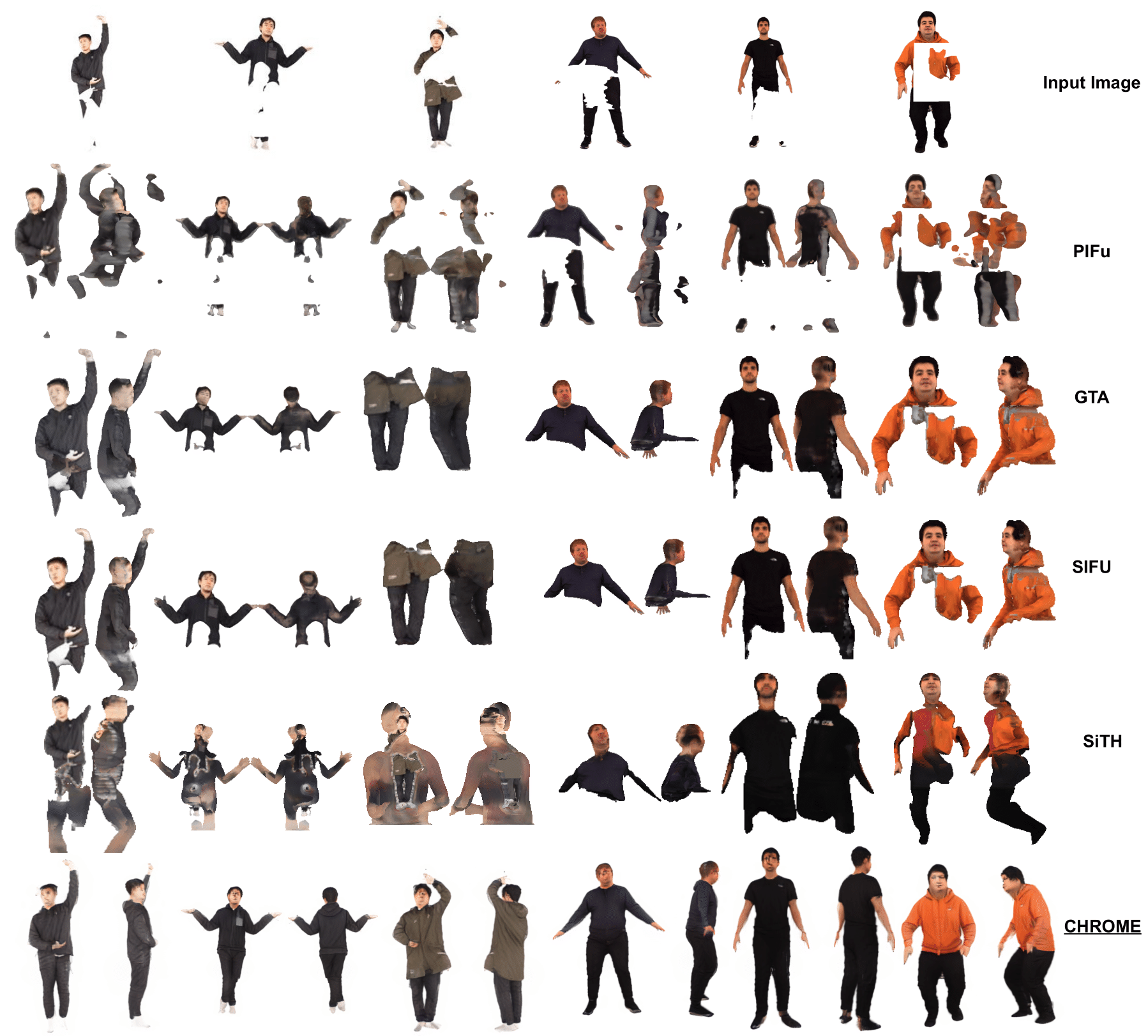}
    \caption{\textbf{Qualitative analysis of $\ours$ on artificially occluded THuman2.0 and CustomHumans:} Qualitative results of $\ours$ against state-of-the-art algorithms~\cite{saito2019pifu, zhang2024global, zhang2024sifu, ho2024sith} on artificially occluded THuman2.0 (top three rows) and  artificially occluded CustomHumans (bottom three rows); clearly none of the baseline algorithm perform occlusion-free novel view synthesis whereas $\ours$ effectively mitigates occlusions and predicts multiview consistent reconstructions. Additional qualitative results are provided in the supplementary. 
    }
    \label{fig:th_ch}
    \vspace{-0.5cm}
\end{figure*}

\section{Experiments and Results}

In this section, we demonstrate \ours's outstanding ability to perform occlusion-free novel view synthesis and geometric reconstruction for single-view occluded human image on multiple datasets, presenting both qualitative and quantitative analyses on multiple benchmarks. Detailed implementation details and more extensive results \eg computation time cost, and flexibility with single/stereo-view as input can be found in supplementary material. 

\noindent \textbf{Datasets and Metrics:}  We evaluate our method on multiple datasets: THuman2.0~\cite{tao2021function4d} (526 high-quality scans of 150 clothing styles across 500 training and 26 evaluation subjects) using a perspective camera (49.1$^\circ$ FOV) with 16 views per subject; CustomHumans~\cite{ho2023learning} containing 600 scans of 80 subjects with 120 garments, augmented with artificial occlusions for zero-shot testing; CAPE~\cite{ma2020cape} containing 15 subjects in tight clothing, modified with synthetic occlusions for zero-shot evaluation, AHP~\cite{zhou2021human} featuring naturally occluded in-the-wild images used for input-view de-occlusion assessment and MultiHuman~\cite{zheng2021deepmulticap} and OCHuman~\cite{zhang2019pose2seg} containing multiple people in regular interactions. We follow~\cite{zhang2024sifu, saito2019pifu, zhang2024global} and use PSNR, SSIM~\cite{wang2004image}, and LPIPS~\cite{zhang2018unreasonable} to evaluate reconstruction accuracy in 2D and chamfer distance (CD) in cm, point-to-surface (P2S) error (cm), and normal consistency (NC)~\cite{ho2024sith} to evaluate 3D reconstruction quality.

\begin{table}[h!]
\centering
\vspace{-1em}
\caption{Quantitative comparison for Novel View Texture Reconstruction on Occluded THuman2.0~\cite{tao2021function4d}. ``SMPL" denotes requiring ground-truth SMPL annotation for model training. ``3D Scan" denotes requiring ground-truth scan-level supervision for model training.} 
\label{tab:thuman-ocl}
\resizebox{0.95\columnwidth}{!}{%
\begin{tabular}{l|c|ccc}
\toprule
\textbf{Algorithm} & \begin{tabular}{@{}c@{}}SMPL \\ /3D Scan\end{tabular} & PSNR $\uparrow$ & SSIM $\uparrow$ & LPIPS $\downarrow$ \\
\midrule
PIFu~\cite{saito2019pifu} & \cmark & 17.11 & .8831  & .1313 \\
GTA~\cite{zhang2024global} & \cmark & 16.27 & .8810 & .1379 \\
SIFU~\cite{zhang2024sifu} & \cmark & 16.19 & .8783 & .1380\\
SiTH~\cite{ho2024sith} & \cmark & 15.98 & .8779 & .1383\\
\midrule
$\ours$ & \xmark & \textbf{20.54} & \textbf{.9098}  & \textbf{.0893}  \\
\bottomrule
\end{tabular}
}
\end{table}

\begin{table}[h!]
\centering
\caption{Quantitative comparison for Geometric Reconstruction on Occluded THuman2.0~\cite{tao2021function4d}.}
\label{tab:thuman-ocl-geo}
\resizebox{0.85\columnwidth}{!}{%
\begin{tabular}{l|c|ccc}
\toprule
\textbf{Algorithm} & \begin{tabular}{@{}c@{}}SMPL \\ /3D Scan\end{tabular} & CD $\downarrow$ & P2S $\downarrow$ & NC $\uparrow$ \\
\midrule
PIFu~\cite{saito2019pifu} & \cmark & 2.744 & 1.766  & 0.582 \\
GTA~\cite{zhang2024global} & \cmark & 3.266 & 1.675 & 0.603 \\
ECON~\cite{xiu2023econ} & \cmark & 3.296 & 1.691 & 0.606 \\
SIFU~\cite{zhang2024sifu} & \cmark & 3.253 & 1.651 & 0.603\\
SiTH~\cite{ho2024sith} & \cmark & 2.748 & 1.746 & 0.592\\
\midrule
$\ours$ & \xmark & \textbf{1.712} & \textbf{1.347}  & \textbf{0.697}  \\
\bottomrule
\end{tabular}
}
\end{table}

\noindent \textbf{Quantitative Results:} We present a quantitative evaluation of our proposed algorithm, $\ours$, against state-of-the-art methods on the artificially occluded THuman2.0 dataset (Table~\ref{tab:thuman-ocl}), assessing novel view reconstruction quality across 16 viewpoints.
$\ours$ achieves statistically significant improvements over existing approaches~\cite{saito2019pifu, zhang2024sifu, zhang2024global}, which exhibit inconsistent de-occlusion performance, with a notable $\approx$3 dB PSNR gain over PIFu~\cite{saito2019pifu}. Further geometric analysis (Table~\ref{tab:thuman-ocl-geo}) 
demonstrates $\ours$'s superiority, particularly in preserving surface details under occlusion, with marked margins across all metrics. These results validate $\ours$'s robustness in addressing occlusions while maintaining geometric fidelity, outperforming baselines for both photometric and geometric reconstruction.


\begin{table}[h!]
\centering
\caption{Quantitative comparison for zero-shot novel view texture reconstruction on Occluded CustomHumans~\cite{ho2023learning}.}
\label{tab:ch-ocl}
\resizebox{0.95\columnwidth}{!}{%
\begin{tabular}{l|c|ccc}
\toprule
\textbf{Algorithm} & \begin{tabular}{@{}c@{}}SMPL \\ /3D Scan\end{tabular} & PSNR $\uparrow$ & SSIM $\uparrow$ & LPIPS $\downarrow$ \\
\midrule
PIFu~\cite{saito2019pifu} & \cmark & 14.77 & .8779 & .1353\\
GTA~\cite{zhang2024global} & \cmark & 13.90 & .8955 & .1274 \\
SIFU~\cite{zhang2024sifu} & \cmark & 13.93 & .8939 & .1273 \\
SiTH~\cite{ho2024sith} & \cmark & 13.87 & 0.8959 & .1284 \\
\midrule
$\ours$ & \xmark & \textbf{18.54} & \textbf{.9130} & \textbf{.0850} \\
\bottomrule
\end{tabular}
}
\end{table}

\begin{table}[h!]
\centering
\vspace{-1em}
\caption{Quantitative comparison for zero-shot Geometric Reconstruction on Occluded CustomHumans~\cite{ho2023learning}.}
\label{tab:ch-ocl-geo}
\resizebox{0.9\columnwidth}{!}{%
\begin{tabular}{l|c|ccc}
\toprule
\textbf{Algorithm} & \begin{tabular}{@{}c@{}}SMPL \\ /3D Scan\end{tabular} & CD $\downarrow$ & P2S $\downarrow$ & NC $\uparrow$ \\
\midrule
PIFu~\cite{saito2019pifu} & \cmark & 5.751 & 3.679 & 0.621 \\
GTA~\cite{zhang2024global} & \cmark & 8.418 & 2.763 & 0.688 \\
ECON~\cite{xiu2023econ} & \cmark & 8.065 & 2.963 & 0.678 \\
SIFU~\cite{zhang2024sifu} & \cmark & 7.408 & 2.867 & 0.689 \\
SiTH~\cite{ho2024sith} & \cmark & 7.783 & 2.985 & 0.675 \\
\midrule
$\ours$ & \xmark & \textbf{4.142} & \textbf{2.392} & \textbf{0.748}\\
\bottomrule
\end{tabular}
}
\end{table}


\begin{table}[h!]
\centering
\caption{Quantitative comparison for zero-shot Input View Texture Reconstruction on AHP~\cite{zhou2021human}.}
\label{tab:ahp}
\resizebox{0.95\columnwidth}{!}{%
\begin{tabular}{l|c|ccc}
\toprule
\textbf{Algorithm} & \begin{tabular}{@{}c@{}}SMPL \\ /3D Scan\end{tabular} & PSNR $\uparrow$ & SSIM $\uparrow$ & LPIPS $\downarrow$ \\
\midrule
PIFu~\cite{saito2019pifu} & \cmark & 14.88 & .8317 & .1620 \\
GTA~\cite{zhang2024global} & \cmark & 13.64 & .8234 & .1754 \\
SIFU~\cite{zhang2024sifu} & \cmark & 13.71 & .8248 & .1727 \\
SiTH~\cite{ho2024sith} & \cmark & 13.44 & .8336 & .1706 \\
\midrule
$\ours$ & \xmark & \textbf{17.77} & \textbf{.8558} & \textbf{.1296} \\
\bottomrule
\end{tabular}
}
\vspace{-1em}
\end{table}

In Tables~\ref{tab:ch-ocl} and~\ref{tab:ch-ocl-geo}, we present \textbf{zero-shot generalization} ability of $\ours$ on artificially occluded CustomHumans~\cite{ho2023learning} dataset in terms of occlusion-free novel view synthesis across 16 views and geometric reconstruction. Clearly, $\ours$ comfortably outperforms all baseline algorithms~\cite{saito2019pifu, zhang2024global, xiu2023econ, zhang2024sifu, ho2024sith}. Finally, in Table~\ref{tab:ahp}, we present the \textbf{zero-shot generalization} capacity of $\ours$ on the AHP~\cite{zhou2021human} dataset, where $\ours$ demonstrates superior performance compared to existing baselines. For the naturally occluded AHP dataset, due to the absence of multi-view ground truths, we evaluate results for input-view de-occlusion. In all cases, we observe that the zero-shot generalization performance of $\ours$ is better than existing algorithms~\cite{saito2019pifu, zhang2024global, zhang2024sifu, ho2024sith, xiu2023econ}.

\begin{figure}[!htb]
    \centering
    \includegraphics[width=1\linewidth]{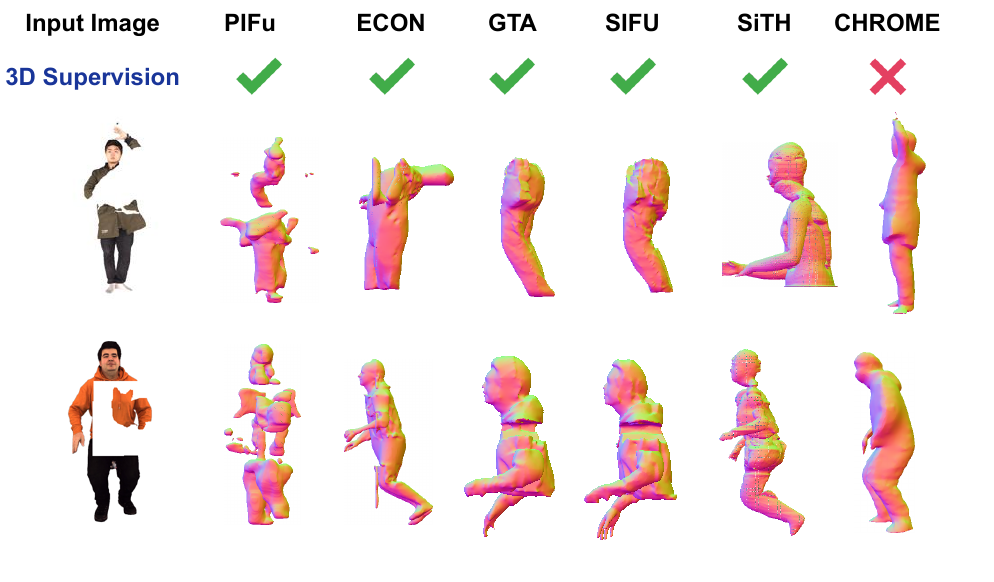}
    \vspace{-1em}
    \caption{\textbf{Qualitative analysis of geometric reconstruction via normal maps:} Qualitative comparisons of $\ours$ against state-of-the-art methods for geometric reconstruction via normal consistency. Clearly, the predictions from existing algorithms are not occlusion-resilient, generating implausible reconstructions whereas $\ours$ effectively handles occlusions, producing occlusion-resilient reconstructions. 
    }
    \label{fig:normal_main}
    \vspace{-1em}
\end{figure}

\begin{figure}[t]
    \centering
    \includegraphics[width=0.8\linewidth]{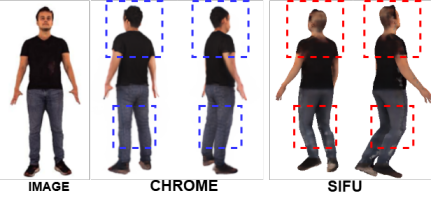}
    \vspace{-1em}
    \caption{\textbf{Clean images:} Qualitative Results on occlusion-free images for \ours~against SIFU.}
    \label{fig:ocl_free}
    \vspace{-1em}
\end{figure}

\begin{figure}[h!]
    \centering
    \includegraphics[width=0.9\linewidth]{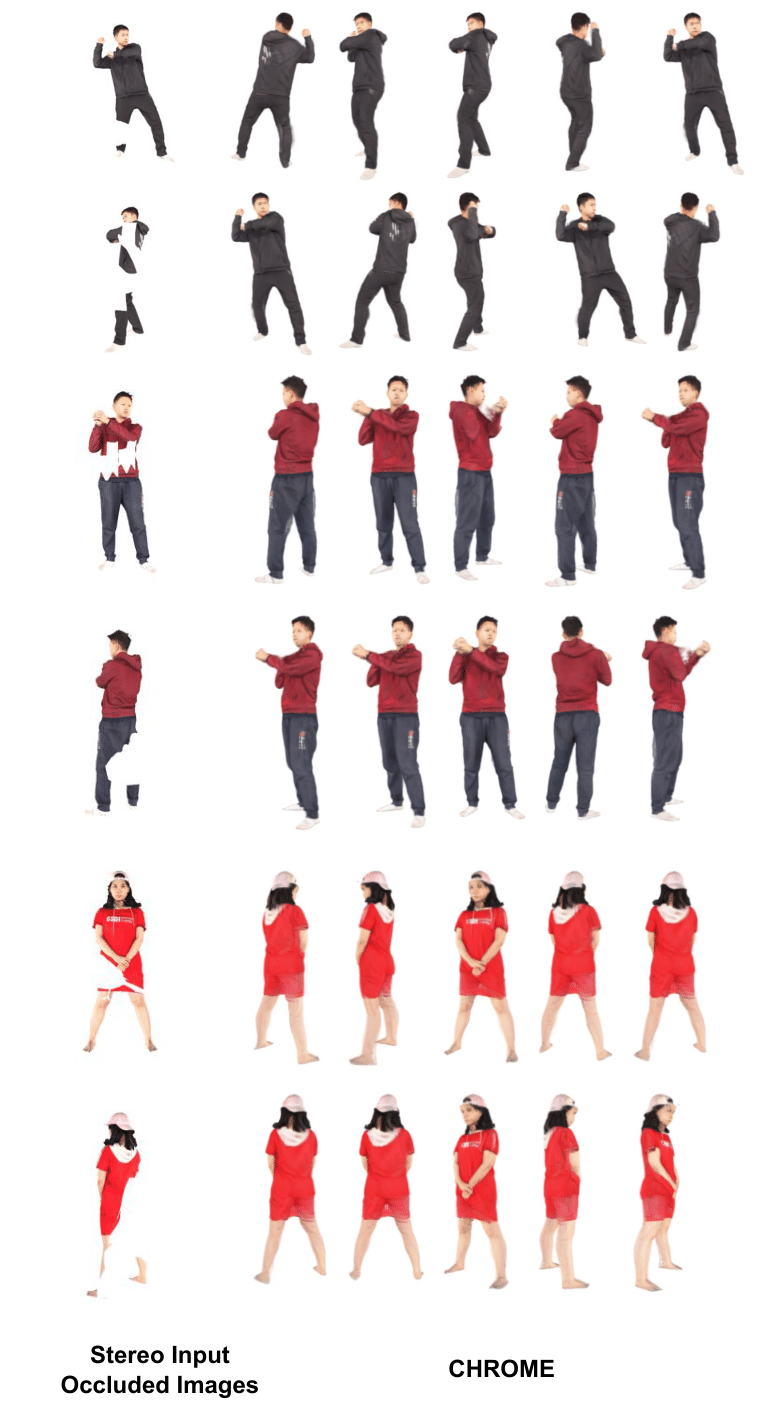}
    \caption{\textbf{Stereo Reconstruction:} Visual outcomes of $\ours$ utilizing stereo occluded images of the same human. 
    }
    \label{fig:stereo_main}
    \vspace{-2em}
\end{figure}

\noindent \textbf{Qualitative Results:} We present qualitative comparisons of $\ours$ with the baseline algorithms in Figures~\ref{fig:th_ch} and \ref{fig:normal_main} in the in-domain artificially occluded datasets THuman2.0~\cite{tao2021function4d}, out-of-domain artificially occluded CustomHumans~\cite{ho2023learning} for novel view synthesis and geometric reconstructions via normal maps. These results clearly show that existing algorithms fail to generate occlusion-free novel views from a single occluded image. Moreover, under occlusions, the generated novel views often suffer from anatomical implausibility and severe multiview inconsistencies. In contrast, $\ours$ consistently produces novel views that are resilient to occlusions and consistent across multiple views. Figure~\ref{fig:ocl_free} shows results of \ours~against SIFU~\cite{zhang2024sifu} on clean, occlusion-free images wherein clearly \ours~outperforms SIFU. Figure~\ref{fig:stereo_main} shows the qualitative results of $\ours$ when stereo inputs are available, showcasing the inherent flexibility of $\ours$ in generating occlusion-free novel views under stereo reconstruction settings. Additional qualitative results are provided in the supplementary. \\


\noindent \textbf{Ablation Study:} Table~\ref{tab:abl} highlights the importance of incorporating the estimated 3D pose~\cite{goel2023humans} from the occluded image as an explicit guidance signal for our multiview diffusion model, $\mathcal{F}_{D}$, on the occluded THuman2.0 dataset~\cite{tao2021function4d}. In particular, even without using the estimated 3D pose as conditioning information, $\ours$ outperforms existing baselines as shown in Table~\ref{tab:thuman-ocl}. However, integrating the pose as explicit guidance leads to more anatomically accurate reconstructions, as shown in Figure~\ref{fig:abl} and discussed in the supplementary.

\begin{table}[h!]
\centering
\caption{Ablation study on the effect of using estimated 3D pose~\cite{goel2023humans} as an explicit pose conditioning on occluded Thuman2.0. Note that, even without using any pose guidance, we outperform baseline algorithms in Table~\ref{tab:thuman-ocl}.}
\label{tab:abl}
\resizebox{0.95\columnwidth}{!}{%
\begin{tabular}{l|ccc}
\toprule
\textbf{Method} & \textbf{PSNR $\uparrow$} & SSIM$\uparrow$ & LPIPS $\downarrow$ \\
\midrule
$\ours$ w/o pose cond. & 20.40 & .8833 & .0966 \\
$\ours$ & \textbf{20.54} & \textbf{.9098} & \textbf{.0893} \\
\bottomrule
\end{tabular}%
}
\end{table}

\begin{figure}[h!]
    \vspace{-2em}
    \centering
    \includegraphics[width=0.9\linewidth]{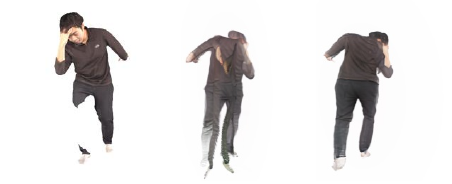}
    \caption{\textbf{Importance of pose conditioning:} An example case where explicit pose guidance enhances anatomical accuracy and visual quality in novel view reconstructions. Left: input occluded image. Middle: novel view synthesis by $\ours$ without pose conditioning. Right: Novel view synthesis by $\ours$.} 
    \label{fig:abl}
    \vspace{-1em}
\end{figure}

\section{Conclusion}
\vspace{-0.5em}

We introduce $\ours$, a novel two-stage algorithm designed for occlusion-resistant, multiview-consistent 3D reconstruction of clothed humans from a single occluded image. Using a pose-controlled diffusion model to generate occlusion-free images and a 3D reconstruction model conditioned on both the initial occluded image and the synthesized multiview images, $\ours$ effectively captures and integrates geometric and texture details across multiple views. This approach ensures robust multiview reconstruction under occlusions, overcoming the limitations of existing methods. The experimental results on challenging datasets demonstrate the superiority of $\ours$ in producing novel view synthesis and geometric reconstructions that are both occlusion resilient and multiview consistent, marking a significant advancement in the field of 3D reconstruction from occluded images. 
\newpage
\noindent \textbf{Acknowledgments.} This work was partially supported by NSF grants CMMI-2326309 and CNS-2312395.


{\small
\bibliographystyle{unsrt}
\bibliography{main}
}

\newpage


\noindent This supplementary material provides additional details and analyses of $\ours$ to complement the main paper. It begins with a discussion of implementation details, including training strategies, hyperparameters, and inference settings. Extended quantitative and qualitative results are presented to evaluate $\ours$ on scenarios such as stereo reconstruction and occlusion-resilient novel view synthesis and geometric reconstruction from single-view images. Also, we analyze the impact of pose estimation on multiview consistency and demonstrate how $\ours$ maintains robustness even with inaccurate pose inputs. We also provide comparative evaluations against existing large reconstruction models, showcasing the superior performance and generalizability of $\ours$. Finally, an inference time analysis highlights its efficiency, making it suitable for real-world applications. Together, these results reinforce the robustness and versatility of $\ours$ across diverse datasets.

\section{Implementation Details}

For all our experiments, we use the PyTorch coding environment with all models being trained on 4$\times$ A40 GPUs. 

We train \(\mathcal{F_{D}}\) for 100 epochs across all experiments. The training utilizes the AdamW optimizer with a cosine annealing learning rate schedule that peaks at \(7 \times 10^{-5}\) within 1000 warm-up steps, after which we use a constant learning rate of \(5 \times 10^{-6}\). During inference, we employ classifier-free guidance with a guidance scale of 4 and 40 diffusion steps, utilizing an ancestral Euler sampling strategy.

For training $\mathcal{F_{R}}$, we use a learning rate initialized at $4 \times 10^{-4}$, which decays following the Cosine Annealing strategy over 100 epochs. In Equation 4 (see main paper), $\lambda_{1}$ is set to 1.5 and $\lambda_{2}$ is set to 1.

\begin{table}[h!]
\centering
\caption{Quantitative comparison for novel view texture reconstruction on regular THuman2.0~\cite{tao2021function4d}.}
\label{tab:thuman-clean}
\resizebox{0.95\columnwidth}{!}{%
\begin{tabular}{l|c|ccc} 
\toprule
\textbf{Algorithm} & SMPL & PSNR $\uparrow$ & SSIM  $\uparrow$ & LPIPS  $\downarrow$ \\
\midrule
SiTH~\cite{ho2024sith} & \cmark & 17.12 & 0.843 & 0.155 \\
GTA~\cite{zhang2024global} & \cmark & 18.05 & - & - \\
SIFU~\cite{zhang2024sifu} & \cmark & \textbf{22.10} & \textbf{.9230} & \textbf{.0790}\\
\midrule
HSGD~\cite{albahar2023single} & \xmark & 17.37 & .8950 & .1300 \\
PIFu~\cite{saito2019pifu} & \xmark & 18.09 & .9110 & .1370 \\
LGM~\cite{tang2024lgm} & \xmark & 20.01 & .8930 & .1160 \\

M123~\cite{xu2024magicanimate} & \xmark & 14.50 & .8740 & .1450 \\

\midrule
$\ours$ & \xmark & \textbf{20.80} & \textbf{.9114}  & \textbf{.0878} \\
\bottomrule
\end{tabular}
}
\end{table}

\begin{figure*}[!htb]
    \centering
    \includegraphics[width=0.75\linewidth]{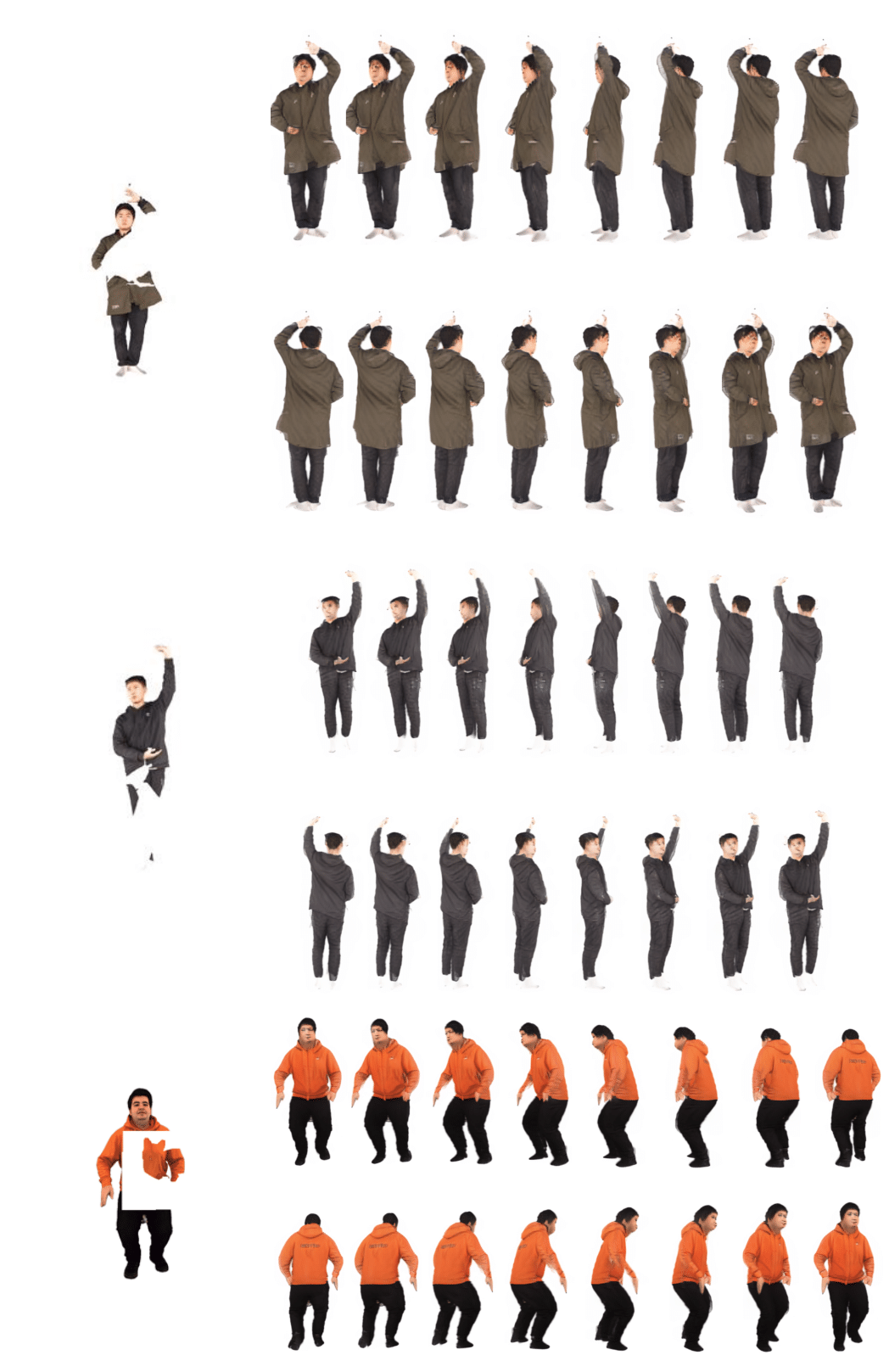}
    \caption{\textbf{Visual Results of $\ours$ on occluded THuman2.0 and CustomHumans:} Qualitative results of $\ours$ on occluded THuman2.0 and occluded CustomHumans.}
    \label{fig:th_ch_supp}
\end{figure*}

\begin{figure*}
    \centering
    \includegraphics[width=1\linewidth]{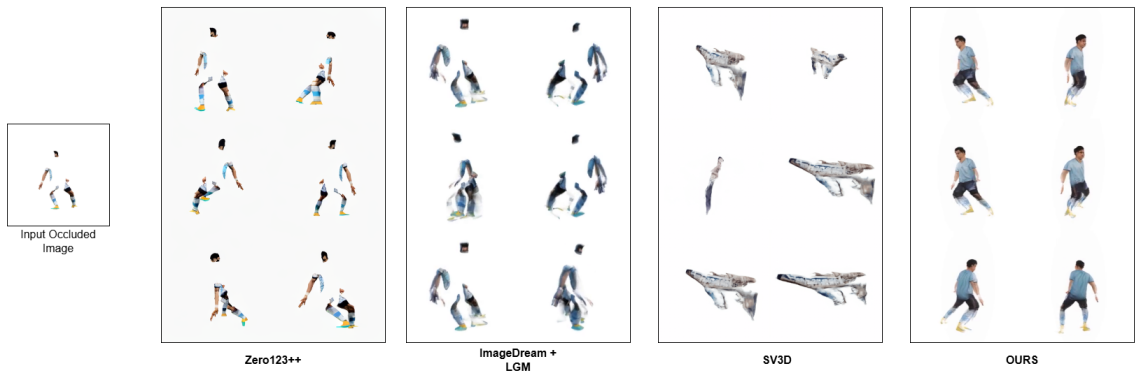}
    \caption{Comparative Qualitative Analysis of Existing Large Reconstruction Models for Zero-Shot Novel View Texture Reconstruction from Occluded Single View Images.}
    \label{fig:lrm}
\end{figure*}

\section{How Robust are Existing Large Reconstruction Models for Occlusion-Free Novel View Synthesis?}

We showcase qualitative comparisons with three baseline algorithms: Zero123++~\cite{shi2023zero123++}, ImageDream with LGM~\cite{tang2024lgm}, and SV3D~\cite{voleti2025sv3d}, utilizing their pretrained weights for the task of occlusion-free novel view reconstruction in Figure~\ref{fig:lrm}. The results reveal significant inconsistencies in these existing methods when applied to our task, underscoring the necessity of a specialized algorithm, $\ours$.

\section{Additional Quantitative Results}

\noindent \textbf{Clean THuman2.0:} In Table~\ref{tab:thuman-clean}, we evaluate the novel-view reconstruction capabilities of $\ours$ under standard occlusion-free conditions. The results show that $\ours$ outperforms all existing methods~\cite{saito2019pifu, tang2024lgm, zhang2024global, albahar2023single, qian2023magic123}, except SIFU~\cite{zhang2024sifu}, for novel view synthesis across 16 views. However, it is important to note that SIFU relies on utilizing SMPL priors and 3D supervision, which may not be available in real-world scenarios. Furthermore, SIFU performs per-subject optimization, taking $\approx$ 6 minutes of computation time per image, while $\ours$ achieves comparable results in $\approx$ 15 seconds. \\

\begin{table}[h!]
\centering
\caption{Quantitative comparison for zero-shot novel view texture reconstruction on Occluded CAPE.}
\label{tab:cape-ocl}
\resizebox{0.95\columnwidth}{!}{%
\begin{tabular}{l|c|ccc}
\toprule
\textbf{Algorithm} & SMPL & PSNR $\uparrow$ & SSIM $\uparrow$ & LPIPS $\downarrow$ \\
\midrule
PIFu~\cite{saito2019pifu} & \xmark & 14.77 & .8779 & .1353\\
GTA~\cite{zhang2024global} & \cmark & 13.90 & .8955 & .1274 \\
SIFU~\cite{zhang2024sifu} & \cmark & 13.93 & .8939 & .1273 \\
SiTH~\cite{ho2024sith} & \cmark & 13.28 & .8782 & .1527 \\
\midrule
$\ours$ & \xmark & \textbf{18.54} & \textbf{.9130} & \textbf{.0850} \\
\bottomrule
\end{tabular}
}
\end{table}
\noindent \textbf{Occluded CAPE:} In Table~\ref{tab:cape-ocl}, we present quantitative results against baseline algorithms for novel view synthesis on occluded CAPE wherein $\ours$ successfully outperforms all baseline algorithms. \\

\begin{table}[h!]
\centering
\small
\begin{tabular}{cccc}
\toprule
\textbf{Methods} & \textbf{PSNR } $\uparrow$ & \textbf{SSIM} $\uparrow$ & \textbf{LPIPS} $\downarrow$\\
\midrule
SD-XL+SIFU~\cite{zhang2024sifu} & 16.27 & .8649 & .1525 \\
\textbf{\ours} & \textbf{20.54} & \textbf{.9098}  & \textbf{.0893} \\
\bottomrule
\end{tabular}
\caption{Novel View Synthesis (NVS) using Inpainting for De-occlusion on Occluded THuman2.0.}
\label{tab:inpaint}
\end{table}

\begin{table}[h!]
    \centering
    \begin{tabular}{l c c}
    \toprule
    \textbf{Occl.} & \textbf{SIFU} & \textbf{CHROME} \\
    \midrule
    25\% & 15.39/.877/.110 & 19.51/.909/.090 \\
    50\% & 14.62/.882/.115 & 19.27/.907/.092 \\
    75\% & 14.04/.880/.123 & 19.06/.904/.094 \\
    \bottomrule
    \end{tabular}
    \caption{Sensitivity to Occlusion Sizes.}
    \label{tab:ocl_size}
\end{table}

\begin{figure}[t]
    \centering
    \includegraphics[width=1\linewidth]{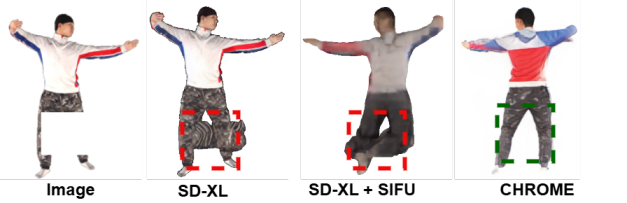}
    \caption{SD-XL + SIFU vs \ours~(zoom in on limbs).}
    \label{fig:sd-inpaint}
\end{figure}

\begin{figure}[t]
    \centering
    \includegraphics[width=1\linewidth]{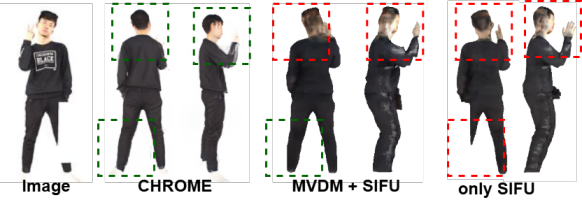}
    \caption{MVDM + SIFU vs \ours.}
    \label{fig:mvdm_sifu}
\end{figure}

\noindent \textbf{Inpainting before Reconstruction:} While a potential solution to our problem could be using a foundational inpainting model such as Stable Diffusion Inpainting, we observe that this model is hardly able to retain the texture or anatomy of the human leading to far from accurate 3D reconstruction, as quantitatively shown in Table~\ref{tab:inpaint} and qualitatively in Figure~\ref{fig:sd-inpaint}. \\

\noindent \textbf{MVDM w/ Existing Algorithms:} We evaluate SIFU on deoccluded images generated by our MVDM ($\mathcal{F_{D}}$) and show its qualitative performance in Fig.~\ref{fig:mvdm_sifu}). \\

\noindent \textbf{Sensitivity to Occlusion levels:} To assess sensitivity to changing occlusion sizes, we evaluated performance at three levels of occlusion (Table~\ref{tab:ocl_size}; where we find that \ours~maintains consistent reconstruction quality even as occlusion severity increases.

\section{$\ours$ for Stereo Reconstruction}

As detailed in the main paper (Section 3), our method, $\ours$, can be seamlessly adapted to stereo reconstruction scenarios, demonstrating its versatility. In Table~\ref{tab:stereo}, we provide comprehensive quantitative results that demonstrate the effectiveness of $\ours$ in stereo reconstruction, \ie, when using two input views. These results underscore the flexibility and robustness of $\ours$ in handling stereo data and achieving high-quality occlusion-resilient reconstructions. Note that, $\ours$ can be trivially extended to handle as many views as the user would like and is upper bounded only by hardware constraints.



\begin{table}[h!]
\centering
\caption{Quantitative comparison of Novel View Texture Reconstruction given stereo inputs on occluded THuman2.0, where the angle represents the separation between the two views relative to the first frame, which is front-facing to the camera.} 
\label{tab:stereo}
\begin{tabular}{cccc}
\toprule
\textbf{Stereo Angle} & \textbf{PSNR $\uparrow$} & \textbf{SSIM $\uparrow$} & \textbf{LPIPS $\downarrow$} \\
\midrule
$45^{\circ}$ & 24.32 & .9280 & .0542 \\
$90^{\circ}$ & 24.70 & .9310 & .0521 \\
$135^{\circ}$ & 24.78 & .9313 & .0511 \\
\bottomrule
\end{tabular}
\end{table}

\begin{table}[h!]
\centering
\caption{Analysis of the Inference Time for Baseline Algorithms with respect to $\ours$ on a NVIDIA A40 GPU}
\label{tab:time}
\resizebox{0.7\columnwidth}{!}{%
\begin{tabular}{cc}
\toprule
\textbf{Algorithm} & Inference Time (Seconds) $\downarrow$ \\
\midrule
PIFu & 33 \\
GTA & 57\\
SIFU & 330\\
SiTH & $\approx$ 300\\
\textbf{\ours{}} & \textbf{15}\\
\bottomrule
\end{tabular}
}
\end{table}

\section{Analyzing Inference Time}

In Table~\ref{tab:time}, we present a comparative analysis of the inference time of $\ours$ versus baseline algorithms. The results demonstrate that $\ours$ achieves superior inference time performance, making it more suitable for real-time applications compared to existing algorithms.

\begin{figure}
    \centering
    \includegraphics[width=1\linewidth]{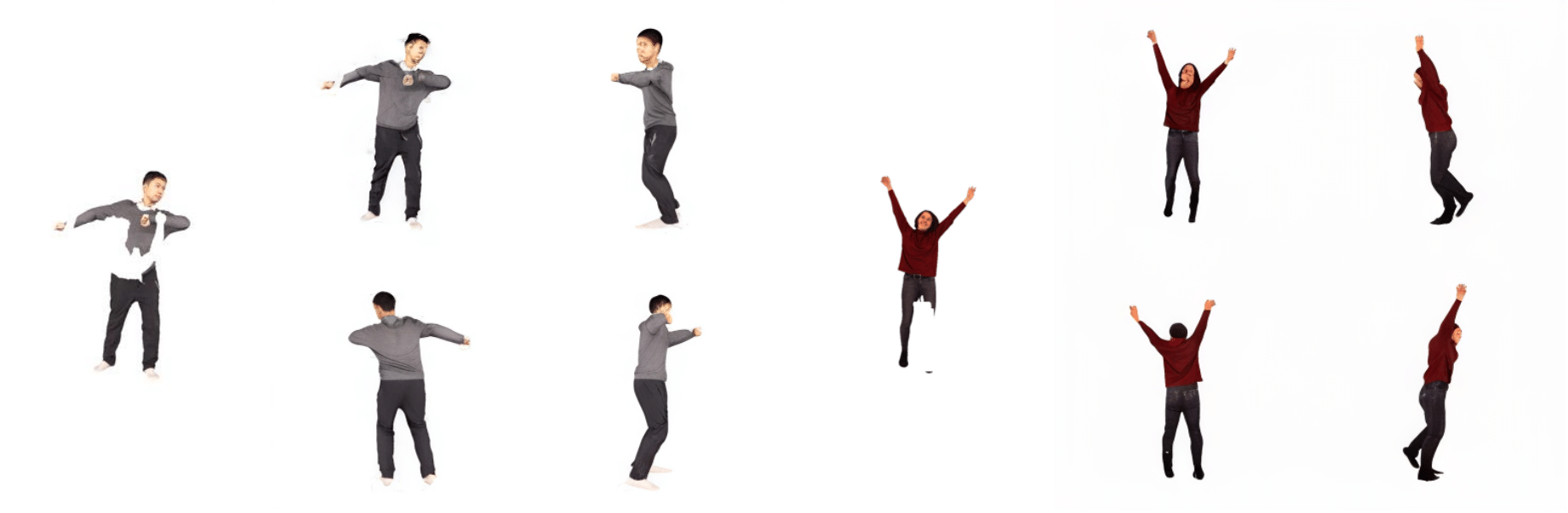}
    \caption{\textbf{Qualitative analysis of Pose Conditioned MVDM:} Qualitative analysis of our ($\mathcal{F}_{D}$) reveals that our pose conditioned MVDM generates reliable occlusion-free images which can later be utilized for 3D reconstruction.}
    \label{fig:diff}

\end{figure}

\begin{figure}[!htb]
    \centering
    \includegraphics[width=1\linewidth]{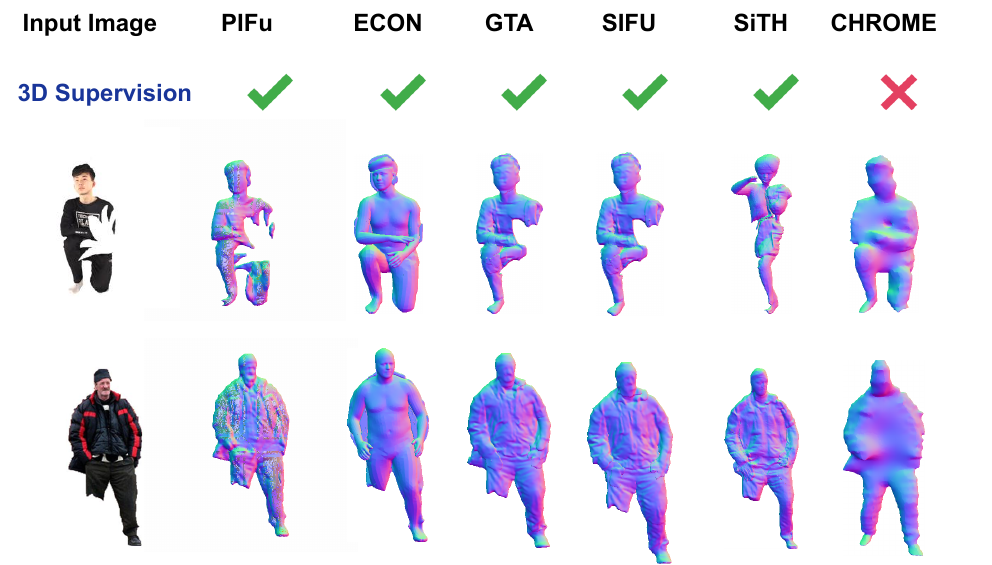}
    \caption{\textbf{Qualitative analysis of geometric reconstruction via normal maps:} Qualitative comparisons of $\ours$ against state-of-the-art methods for geometric reconstruction via normal consistency. Note that $\ours$ does not require 3D mesh supervision during training whereas all baselines necessitate the same.
    }
    \label{fig:normal}
\end{figure}

\begin{figure}[!htb]
    \centering
    \includegraphics[width=1\linewidth]{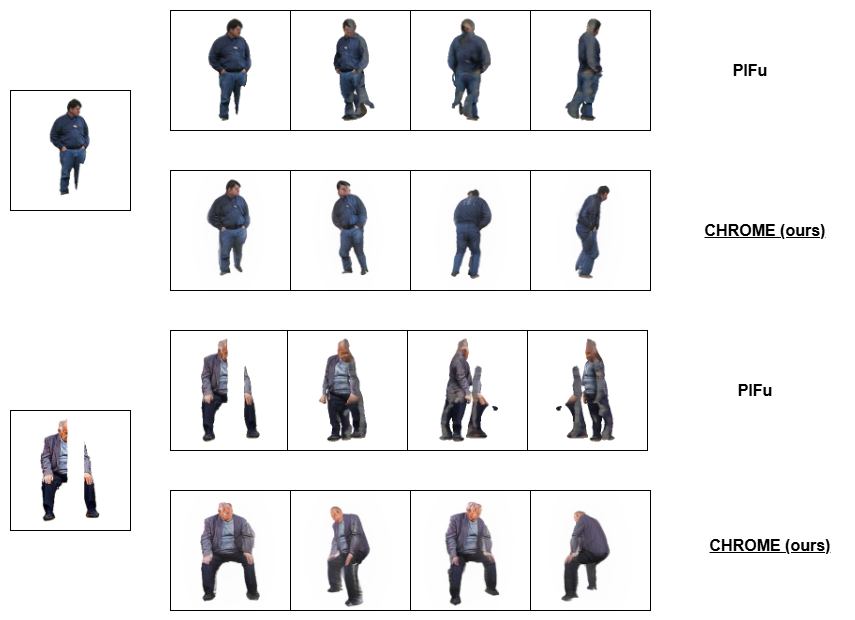}
    \caption{\textbf{Qualitative analysis of $\ours$ on AHP:} Qualitative comparisons of $\ours$ with state-of-the-art method PIFu~\cite{saito2019pifu} on the naturally occluded AHP dataset in zero-shot settings. Clearly, the predictions from PIFu are not occlusion-resilient whereas $\ours$ effectively handles occlusions, producing multiview consistent reconstructions.}
    \label{fig:ahp}
\end{figure}

\begin{figure}[!htb]
    \centering
    \includegraphics[width=1\linewidth]{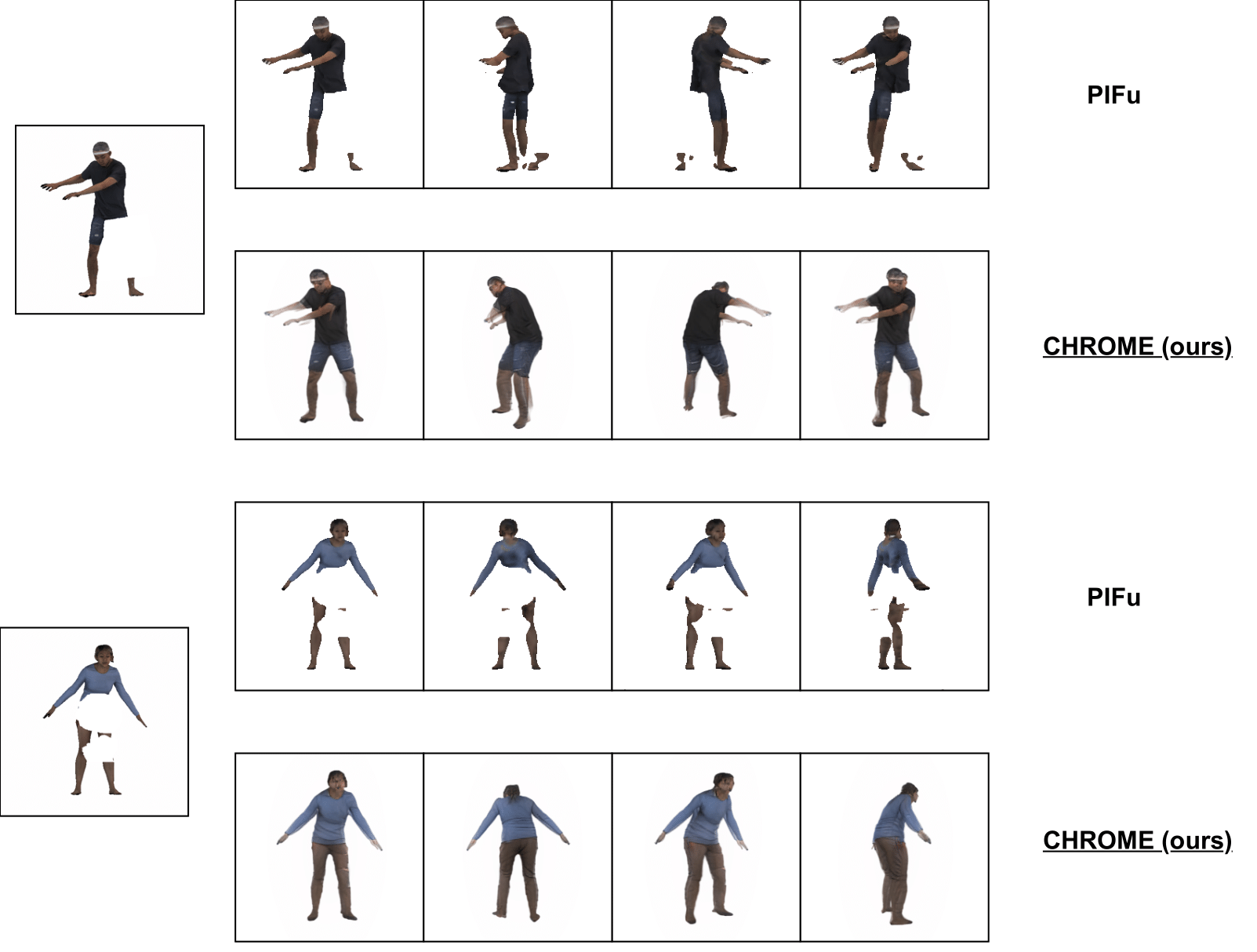}
    \caption{\textbf{Qualitative analysis of $\ours$ on artificially occluded CAPE:} Qualitative comparisons of $\ours$ with state-of-the-art method PIFu~\cite{saito2019pifu} on artificially occluded CAPE in zero-shot settings. Clearly, the predictions from PIFu are not occlusion-resilient whereas $\ours$ effectively handles occlusions, producing multiview consistent reconstructions.}
    \label{fig:cape}
\end{figure}

\begin{figure}
    \centering
    \includegraphics[width=1\linewidth]{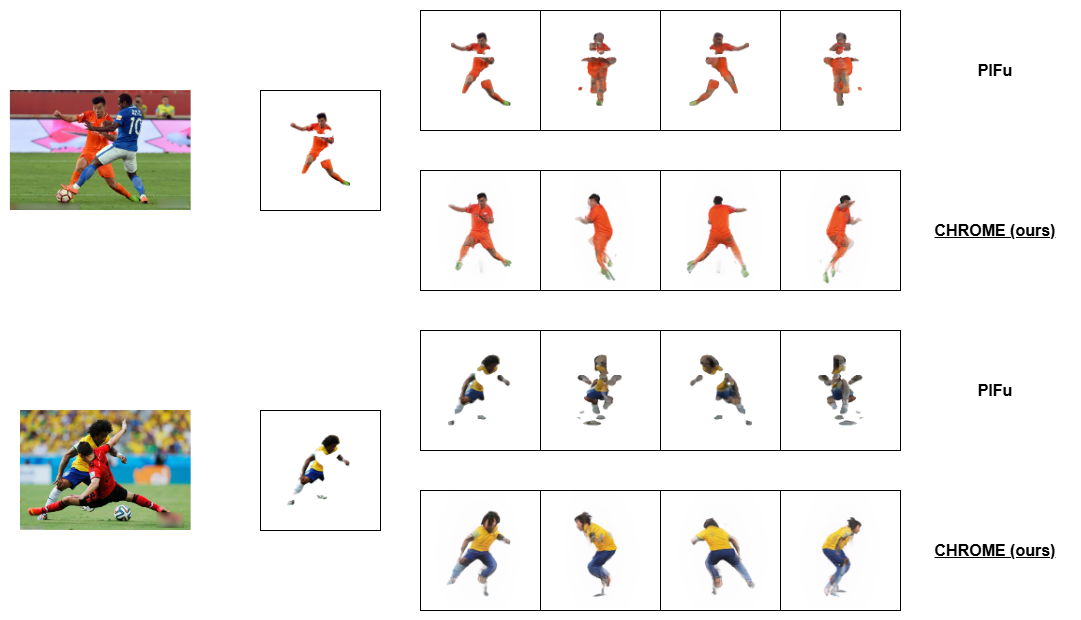}
    \caption{\textbf{Qualitative analysis of $\ours$ on OCHuman:} Qualitative comparisons of $\ours$ with state-of-the-art method PIFu~\cite{saito2019pifu} on the naturally occluded OCHuman dataset in zero-shot settings. Clearly, the predictions from PIFu are not occlusion-resilient whereas $\ours$ effectively handles occlusions, producing multiview consistent reconstructions.}
    \label{fig:och}

\end{figure}

\begin{figure}
    \centering
    \includegraphics[width=1\linewidth]{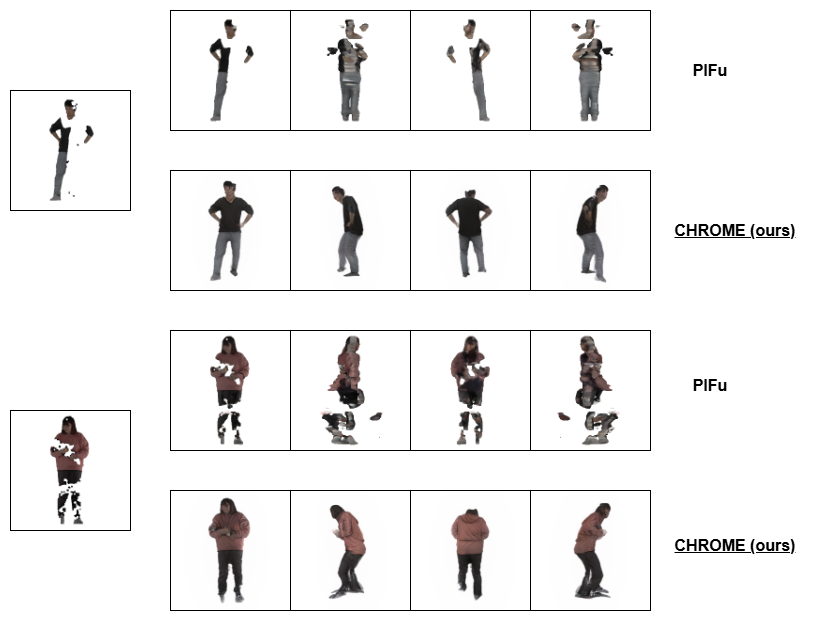}
    \caption{\textbf{Qualitative analysis of $\ours$ on MultiHuman:} Qualitative comparisons of $\ours$ with state-of-the-art method PIFu~\cite{saito2019pifu} on the naturally occluded MultiHuman dataset in zero-shot settings. Clearly, the predictions from PIFu are not occlusion-resilient whereas $\ours$ effectively handles occlusions, producing multiview consistent reconstructions.}
    \label{fig:mh}

\end{figure}

\section{Additional Qualitative Results}

We provide more qualitative results on the occluded THuman2.0 and occluded CustomHumans as an extension of the main paper in Figure~\ref{fig:th_ch_supp}.
We provide a qualitative analysis of $\mathcal{F}_{D}$ for reliable occlusion-free novel view synthesis in Figure~\ref{fig:diff}. We also provide a visualization of normal maps against baselines in Figure~\ref{fig:normal}. Furthermore, we provide qualitative results on the AHP~\cite{zhou2021human}, artificially occluded CAPE~\cite{ma2020cape}, OCHuman~\cite{zhang2019pose2seg} and MultiHuman~\cite{zheng2021deepmulticap} datasets in Figures~\ref{fig:ahp}, \ref{fig:cape}, \ref{fig:och} and~\ref{fig:mh}. AHP, OCHuman, and MultiHuman feature instances of natural occlusions, where existing state-of-the-art (SOTA) algorithms, such as PIFu~\cite{saito2019pifu}, tend to perform poorly. In contrast, $\ours$ shows superior performance by providing high-quality reconstructions that are robust to occlusions. We show qualitative results only against PIFu as we find it to be the best performing baseline algorithm in terms of quantitative performance.

\begin{figure}
    \centering
    \includegraphics[width=1\linewidth]{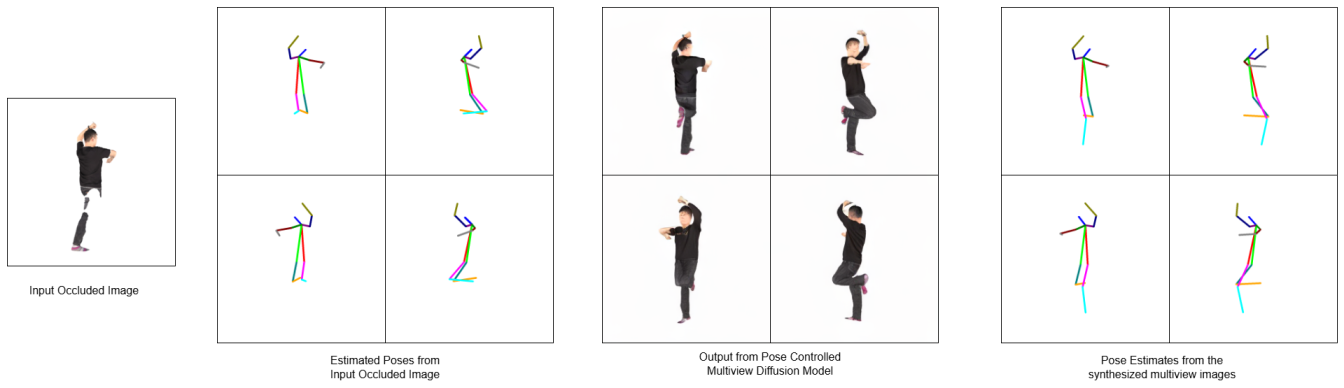}
    \caption{Analyzing the Impact of Pose Estimation on Multiview Reconstruction: Observe the differences between the pose estimates derived from the input occluded image and those obtained from the synthesized multiview images. The conditioning of $\mathcal{F}_{D}$ on the input occluded image ensures that the synthesized images preserve information originating from the input occluded image.}
    \label{fig:pose5}

\end{figure}

\begin{figure}
    \centering
    \includegraphics[width=1\linewidth]{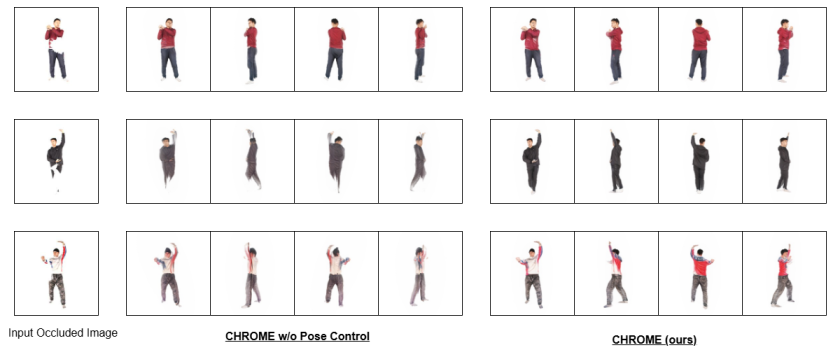}
    \caption{Additional Qualitative Results highlighting the importance of using Pose estimates as explicit control guidance.}
    \label{fig:wopose}
\end{figure}

\section{Analyzing the Effect of Pose Estimator on Multiview Reconstruction}

In Section 3 of the main paper, we highlight that even when pose estimation is inaccurate, the MVDM model ($\mathcal{F}_{D}$) ensures that the generated multiview images remain consistent with the input image. Specifically, the model preserves the visible regions of the input image and reconstructs only the occluded parts based on the provided pose information and the occluded input image itself. This is qualitatively illustrated in Figure~\ref{fig:pose5}, where the 2D projections of the estimated 3D pose are noticeably incorrect and implausible (particularly for the legs). Despite this, $\mathcal{F}_{D}$ successfully generates multiview reconstructions that align with the occluded input image and pose conditioning, producing plausible multiview outputs. Additional qualitative results for novel view synthesis using the $\mathcal{F_{D}}$ trained without incorporating pose information (discussed in Ablation Study, Section 4 of the main paper) is presented in Figure~\ref{fig:wopose}.

\section*{Limitations and Weaknesses}

\noindent \textbf{Limitations:} It should be noted that our solution may suffer from the domain gap between training and inference poses. Prior-based augmentations could be considered in future work, to improve generalizability. 

\noindent \textbf{Societal Impacts:} While $\ours$ may be used for unwanted re-identification and surveillance, we believe that our method can positive impact our community by lowering technical barriers for broader participation, \eg, in VR/AR applications and other creative processes.

\end{document}